\documentclass[10pt,twocolumn,letterpaper]{article}

\usepackage[pagenumbers]{cvpr} 

\usepackage{cvpr}              

\usepackage{graphicx}
\usepackage{placeins}
\usepackage{amsmath}
\usepackage{amssymb}
\usepackage{booktabs}
\usepackage[misc]{ifsym}
\usepackage{times}
\usepackage{epsfig}
\usepackage{multirow}
\usepackage{soul}
\usepackage[accsupp]{axessibility}  
\usepackage{xspace}
\usepackage{mathrsfs}
\usepackage{soul}
\usepackage{array}
\usepackage{tabu}
\usepackage{color,xcolor}
\usepackage{colortbl}
\usepackage{threeparttable}
\definecolor{mygray}{gray}{.85}
\definecolor{mygray1}{gray}{.7}
\definecolor{mygray2}{gray}{.93}

\usepackage{makecell}
\usepackage{flushend}
\newcommand{\mevisdataset}{MeViS\xspace}
\newcommand{\mevisfullname}{\textbf{M}otion \textbf{e}xpressions \textbf{Vi}deo \textbf{S}egmentation\xspace}

\makeatletter
\newcommand{\thickhline}{%
    \noalign {\ifnum 0=`}\fi \hrule height 1pt
    \futurelet \reserved@a \@xhline
}

\usepackage[pagebackref,breaklinks,colorlinks]{hyperref}

\usepackage[capitalize]{cleveref}
\crefname{section}{Sec.}{Secs.}
\Crefname{section}{Section}{Sections}
\Crefname{table}{Table}{Tables}
\crefname{table}{Tab.}{Tabs.}

\def\cvprPaperID{*****} 
\def\confName{CVPR}
\def\confYear{2023}

\newcommand{\teamtitle}[1]{\noindent\textit{\textbf{Title:}} #1 \par}
\newcommand{\teammembers}[1]{\noindent\textit{\textbf{Members:}} #1 \par}
\newcommand{\teamaffiliations}[1]{\noindent\textit{\textbf{Affiliations:}} {\parindent0pt \par #1} \par \vspace{2mm}}

\begin{document}

\title{PVUW 2024 Challenge on Complex Video Understanding:\\ Methods and Results}

\author{Henghui Ding\footnotemark[2]\quad Chang Liu\footnotemark[2]\quad Yunchao Wei\footnotemark[2]\quad Nikhila Ravi\footnotemark[2]\quad Shuting He\footnotemark[2]\quad Song Bai\footnotemark[2]\quad Philip Torr\footnotemark[2] \\
Deshui Miao\quad Xin Li\quad Zhenyu He\quad Yaowei Wang\quad Ming-Hsuan Yang\quad\\
Zhensong Xu\quad Jiangtao Yao\quad Chengjing Wu\quad Ting Liu\quad Luoqi Liu\\
Xinyu Liu\quad Jing Zhang\quad Kexin Zhang\quad Yuting Yang\quad Licheng Jiao\quad Shuyuan Yang\\
Mingqi Gao\quad Jingnan Luo\quad Jinyu Yang\quad Jungong Han\quad Feng Zheng\\
Bin Cao\quad Yisi Zhang\quad Xuanxu Lin\quad Xingjian He\quad Bo Zhao\quad Jing Liu\quad\\
Feiyu Pan\quad Hao Fang\quad Xiankai Lu\\
}
\maketitle
\renewcommand{\thefootnote}{\fnsymbol{footnote}}
\footnotetext[2]{CVPR 2024 PVUW Workshop \& Challenge organizers. All others are challenge participants from the top-3 teams of MOSE and MeViS tracks.}
\footnotetext[0]{${\textrm{\Letter}}$ henghui.ding@gmail.com}

\begin{abstract}
   Pixel-level Video Understanding in the Wild Challenge (PVUW) focus on complex video understanding. In this CVPR 2024 workshop, we add two new tracks, Complex Video Object Segmentation Track based on MOSE dataset and Motion Expression guided Video Segmentation track based on MeViS dataset. In the two new tracks, we provide additional videos and annotations that feature challenging elements, such as the disappearance and reappearance of objects, inconspicuous small objects, heavy occlusions, and crowded environments in MOSE. Moreover, we provide a new motion expression guided video segmentation dataset MeViS to study the natural language-guided video understanding in complex environments. These new videos, sentences, and annotations enable us to foster the development of a more comprehensive and robust pixel-level understanding of video scenes in complex environments and realistic scenarios. The MOSE challenge had 140 registered teams in total, 65 teams participated the validation phase and 12 teams made valid submissions in the final challenge phase. The MeViS challenge had 225 registered teams in total, 50 teams participated the validation phase and 5 teams made valid submissions in the final challenge phase.
\end{abstract}

\section{Introduction}
\label{sec:intro}

Pixel-level Scene Understanding~\cite{li2023transformer,CCL,wu2024towards} is one of the fundamental problems in computer vision, which aims at recognizing object classes, masks and semantics of each pixel in the given image. Since the real-world is actually video-based rather than a static state, learning to perform video segmentation is more reasonable and practical for realistic applications. To advance the segmentation task from images to videos, we will present new datasets and competitions in this workshop, aiming at performing the challenging yet practical Pixel-level Video Understanding in the Wild (PVUW). In this year, we add two new tracks, Complex Video Object Segmentation Track based on MOSE~\cite{MOSE} and Motion Expression guided Video Segmentation track based on MeViS~\cite{MeViS}. In the two new tracks, we provide additional videos and annotations that feature challenging elements, such as the disappearance and reappearance of objects, inconspicuous small objects, heavy occlusions, and crowded environments in MOSE. Moreover, we provide a new motion expression guided video segmentation dataset MeViS to study the natural language-guided video understanding in complex environments.

\begin{figure*}
    \includegraphics[width=0.999\textwidth]{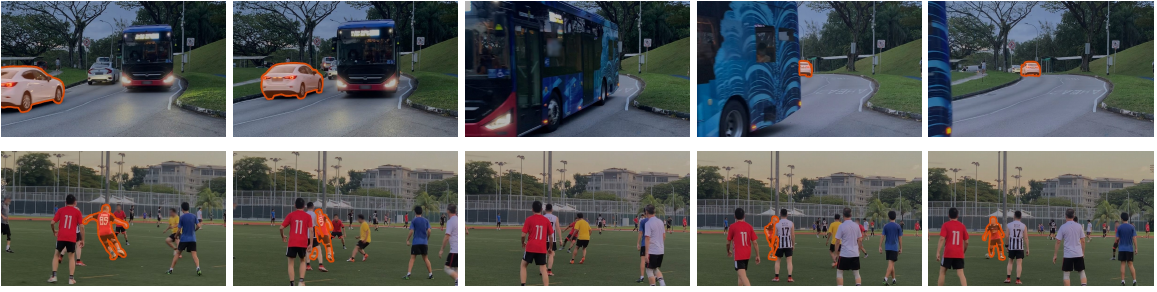}
    \vspace{-7.6mm}
    \caption{Example Videos of co\textbf{M}plex video \textbf{O}bject \textbf{SE}gmentation (\textbf{MOSE}) dataset~\cite{MOSE}. The standout feature of the MOSE dataset is its complex scenes, which include the disappearance and reappearance of objects, small and inconspicuous objects, heavy occlusions, and crowded environments. The aim of the MOSE dataset is to foster the development of complex video understanding.}
\end{figure*}

\begin{figure*}
    \includegraphics[width=0.999\textwidth]{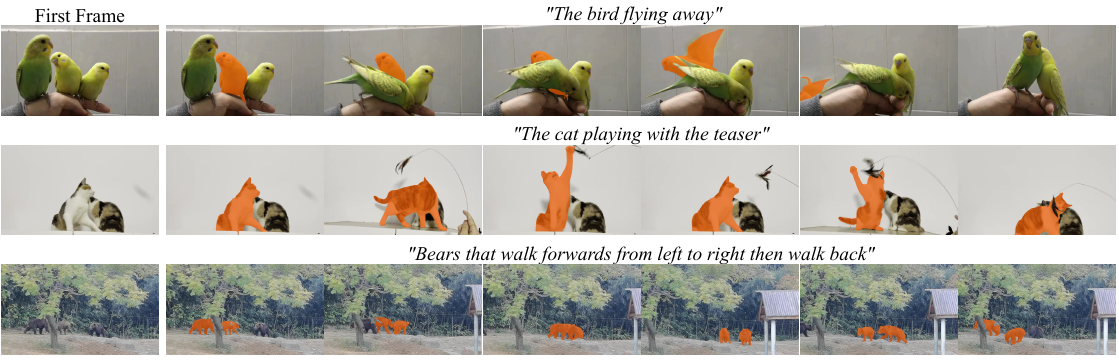}
    \vspace{-7.6mm}
    \caption{Example Videos of \mevisfullname (\textbf{\mevisdataset}) dataset~\cite{MeViS}. The expressions in MeViS mainly emphasize motion attributes, making it impossible to identify the referred target object by looking at a single frame. The aim of the MeViS dataset is to foster the development of motion understanding in complex scenes.}
\end{figure*}
Video object segmentation (VOS) focuses on segmenting specific objects throughout an entire video sequence. While state-of-the-art VOS methods have achieved impressive results (\eg, over 90\% $\mathcal{J}\&\mathcal{F}$) on existing datasets, these datasets typically feature targets that are salient, dominant, and isolated. Consequently, VOS in complex scenes remains underexplored. To address this and enhance real-world applicability, Ding et al.~\cite{MOSE} introduce a new dataset named coMplex video Object SEgmentation (MOSE), designed to study tracking and segmenting objects in complex environments. MOSE's standout feature is the inclusion of crowded and occluded scenes, where target objects often get obstructed or disappear in some frames. The experiments on MOSE demonstrate that current VOS algorithms struggle with object perception in complex scenes. For example, in the semi-supervised VOS setting, the top-performing state-of-the-art method achieves only 59.4\% $\mathcal{J}\&\mathcal{F}$ on MOSE, significantly lower than their ~90\% $\mathcal{J}\&\mathcal{F}$ performance on DAVIS. These findings highlight the unresolved challenges in complex scenes and indicate a need for further research to address these challenges.

Referring Video Object Segmentation (RVOS) focuses on segmenting specific objects throughout an entire video sequence based on sentences describing the target objects. Current referring video object datasets usually emphasize salient objects and include language expressions with many static attributes, allowing target identification in a single frame~\cite{liu2023gres}. These datasets neglect the role of motion in language-guided video object segmentation. To explore the potential of using motion expressions for object segmentation in videos, Ding et al.~\cite{MeViS} introduce a large-scale dataset called MeViS, featuring numerous motion expressions to identify target objects in complex environments. The experiments on MeViS show that current RVOS methods struggle with motion expression-guided segmentation. The image-based referring segmentation methods~\cite{VLTPAMI,ding2021vision,ISFP,MA3Net,ding2020phraseclick} cannot well understand the motion information in videos. These findings highlight the unresolved challenges in motion understanding under complex scenes and indicate a need for further research to address these challenges.

\section{Tracks and Datasets}
\textbf{The MOSE Track} is based on the MOSE dataset~\cite{MOSE}, which focuses on the task of Video Object Segmentation (VOS), especially in real-world complex and dense scenes. The dataset contains 2,149 videos and 5200 objects annotated with 431,725 segmentation masks. The dataset is split into three subsets, including training, validation, and test. The final testing data for the competition is built on partial of the test set. This part of data was private before, and was made open for the first time for the competition. 

One of the most unique features of the dataset is its focus on complex scenes in the task of VOS, such as heavy occlusions, crowded scenarios, and objects that disappear and reappear. It emphasizes the need for stronger association algorithms to track objects with changing appearances and promotes research in occlusion understanding, attention to small and inconspicuous objects, and tracking in crowded environments. The dataset's complexity and length pose significant challenges for current VOS methods, highlighting the need for advancements in complex video object segmentation.

\textbf{The MeViS Track} is based on the newly proposed large-scale motion expression-guided video segmentation dataset, MeViS~\cite{MeViS}. Methods are required to extract and segment the target object based on a expression that describes the motion of the object, in a long video. The dataset is build with 2,006 videos. 8,171 objects are annotated with more than 443,000 segmentation masks and 28,570 motion expressions. The annotation data scale of MeViS is significantly larger than other existing language-guided video segmentation dataset. Similarly with MOSE, partial of the test set of MeViS is used as the testing data for the competition. This part of data is also made public for the first time for the competition. 
 
The dataset focuses on describing the motion of objects in videos through language expressions, emphasizing the significance of temporal properties. It challenges current video object segmentation methods by requiring the identification of objects based solely on their motion, without relying on static attributes like color or category names. MeViS presents a complex environment where multiple objects coexist with motion, making it difficult to identify targets through saliency or category information alone, thus pushing the boundaries of language-guided video understanding in dynamic scenarios.

\textbf{Competition Overview.} Both tracks are hosted on the CodaLab platform~\cite{codalab_competitions_JMLR}. For valid and challenge phase, participants are only given input data, while the full ground-truth are kept private. All participants are required to register on the platform for evaluation. Data for the validation phase is always open for download and evaluation, but the data for the final challenge phase is only available for download and evaluation during the challenge phase of 10 days. The number of submission for each team is not limited for valid phase but is limited to 5 for test phase.

\begin{table}[t]
   \renewcommand\arraystretch{1.2}
   \centering
   \caption{\textbf{MOSE Challenge results and final rankings.}}
   \vspace{-3mm}
   \setlength{\tabcolsep}{2.96mm}{\begin{tabular}{clrrr}
      \thickhline
\rowcolor{gray!60} Rank & Team  & \multicolumn{1}{l}{$\mathcal{J}$} & \multicolumn{1}{l}{$\mathcal{F}$} & \multicolumn{1}{l}{$\mathcal{J\&F}$} \\
     \hline
     1     & PCL\_VisionLab & 81.0  & 87.9  & 84.5 \\
     \rowcolor{gray!20}2     & Yao\_Xu\_MTLab & 80.1  & 86.8  & 83.5 \\
     3     & ISS   & 78.8  & 85.6  & 82.2 \\
     \rowcolor{gray!20}4     & xsong2023 & 78.7  & 85.4  & 82.1 \\
     5     & yangdonghan50 & 78.0  & 84.8  & 81.4 \\
     \rowcolor{gray!20}6     & YongxinWang & 77.2  & 84.0  & 80.6 \\
     7     & Tapallai & 77.0  & 84.0  & 80.5 \\
     \rowcolor{gray!20}8     & guojuan & 74.5  & 81.5  & 78.0 \\
     9     & jmy   & 74.3  & 81.3  & 77.8 \\
     \rowcolor{gray!20}10    & lll7733 & 71.8  & 79.6  & 75.7 \\
     11    & cc886 & 69.1  & 76.6  & 72.8 \\
    \rowcolor{gray!20} -    & (Baseline) & 67.3  & 74.8  & 71.0 \\
     12    & cqbu  & 63.4  & 70.6  & 67.0 \\
     \thickhline
     \end{tabular}}%
   \label{tab:results_mose}%
 \end{table}%
 
To evaluate the performance of methods, both tracks employs the standard and commonly recognized Jaccard ($\mathcal{J}$) metric for region similarity and F-measure ($\mathcal{F}$) for contour accuracy, as evaluation metrics, as in previous works~\cite{MOSE,MeViS,Perazzi2016,vos2018,seo2020urvos,DsHmp}. The average of $\mathcal{J}$ and $\mathcal{F}$ is used as the overall performance of the methods. The final ranking is based on the average of $\mathcal{J}$ and $\mathcal{F}$ (denoted as $\mathcal{J\&F}$) on the test set.


\section{MOSE Challenge Methods and Teams}

\begin{figure*}[htbp]
\centering
\includegraphics[width=0.98\linewidth]{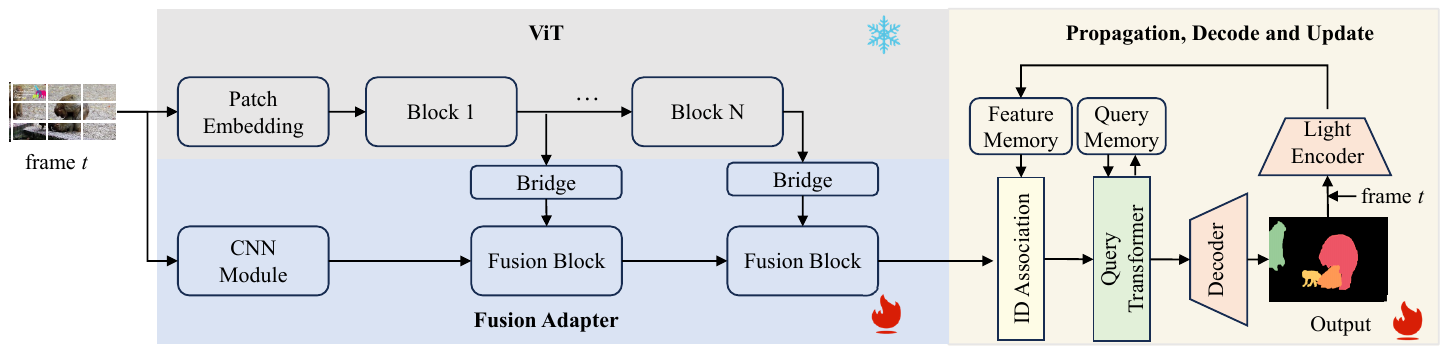}
\caption{Overall framework of PCL\_VisionLab team method, 1st place solution for MOSE Challenge in CVPR 2024.}
\label{fig:overall_pclmds}
\end{figure*} 

For MOSE~\cite{MOSE} track, from 140 teams registered in total, 65 teams participated the validation phase and 12 teams entered the in the challenge phase. The final results of the MOSE Track are reported in \Cref{tab:results_mose}.

\noindent\href{https://henghuiding.github.io/MOSE/ChallengeCVPR2024}{https://henghuiding.github.io/MOSE/ChallengeCVPR2024}

\subsection{PCL\_VisionLab team} \label{sec:pcl}
\teamtitle{1st Place Solution for MOSE Track in CVPR 2024 PVUW Workshop: Complex Video Object Segmentation \cite{miao1st}}
\teammembers{Deshui Miao\textsuperscript{1,}\textsuperscript{2}, Xin Li\textsuperscript{2}, Zhenyu He\textsuperscript{1,}\textsuperscript{2}, Yaowei Wang\textsuperscript{2}, and Ming-Hsuan Yang\textsuperscript{3}}
\teamaffiliations{\textsuperscript{1}Harbin Institute of Technology (ShenZhen)\par
\textsuperscript{2}Peng Cheng Laboratory\par
\textsuperscript{3}University of California at Merced}

To solve the problems of VOS, we propose a robust semantic-aware and query-enhanced video object segmentation method.
In this solution, we first introduce the proposed fusion block, which utilizes the semantic and detailed information of the pretrained ViT models. 
This helps us deal with complex target appearance variance and ID confusion between targets with similar appearances. 
In detail, we fuse the information of the cls token from the ViT to multi-scale features and conduct local fusion between frame patches and multi-scale features for detailed fusion. 
In addition, to ensure the target representation of the target queries, we develop a discriminative query representation module in the query transformer to capture the local representation of the targets.

\subsubsection{Fusion Block}
Since the VOS task involves generic objects without class labels, learning semantic representations directly from the VOS dataset during training is challenging. 
However, the CLS token in a pre-trained ViT captures semantic information from the entire image, providing a comprehensive, global representation of the image content. 
By integrating the CLS token with multi-scale features generated from CNN networks, we can acquire detailed semantic features at various scales.
In Figure~\ref{fig:overall_pclmds}, cross-attention is used to perform semantic prior learning for VOS.

Then, multi-scale deformable cross-attention is utilized to learn the spatial dependence of different scale features, which helps handle objects with complex structures or separate parts.
\begin{figure*}[htbp]
  \centering
  \includegraphics[width=0.8\linewidth]{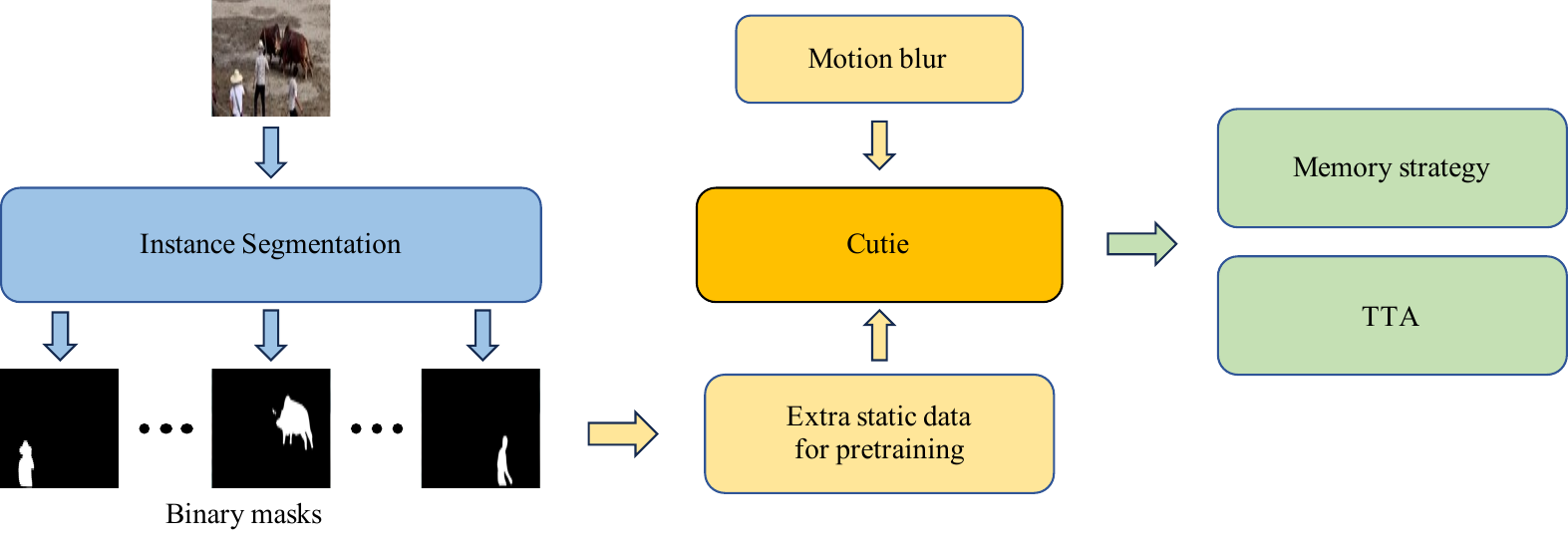}

   \caption{Overall framework of Yao\_Xu\_MTLab team method, 2nd place solution for MOSE Challenge in CVPR 2024.}
   \label{fig:1}
\end{figure*}

\subsubsection{Discriminative Query Generation}
We note that updating the target query memory directly with entire object patches generated based on online predicted masks is ineffective as the predicted masks often cover background noise, reducing target distinctiveness and leading to accumulating errors over time.
To propagate target queries effectively across frames, we update the target queries with the most distinctive feature of the target object.

In detail, we select the discriminative feature of a target object by comparing the target query with every channel activation in the correlated feature map of the target and taking the most similar one.
Based on the discriminative target feature generated from a new target sample, we can update target queries by dynamically calculating the relationship between the salient query and salient pixel features in an additive manner.
The proposed discriminative query generation scheme adaptively refines target queries with the most representative features, which helps deal with the challenges of dramatic appearance variations in long-term videos.

\subsubsection{Experiments}

\noindent \textbf{Training}. 
Our training settings are similar to Cutie's. 
To enhance the performance of our model, we utilize the MEGA dataset constructed by Cutie, which includes the YouTubeVOS, DAVIS, OVIS, MOSE, and BURST datasets.
We sample eight frames to train the model, and three are randomly selected to train the matching process.
For each sequence, we randomly choose at most three targets for training.
The point supervision in loss is adopted to reduce the memory requirements.
We train the model for 195k on the MEGA dataset.
All our models are trained on 8 x NVIDIA V100 GPUs and tested on an NVIDIA V100 GPU.

\noindent\textbf{Inference.} 
Our feature and query memory is updated every 3rd frame during the testing phase. 
For longer sequences, we employ a long-term fusion strategy for updating. To enhance storage quality, we skip frames without targets and do not store them.
The test input size contains two scales: 720 for general size and 1080 for small targets.
The final score is a version of multi-scale fusion.

\noindent \textbf{Evaluation Metrics.} 
We use mean Jaccard $\mathcal{J}$ index and mean boundary
$\mathcal{F}$ score, along with mean $\mathcal{J}\&\mathcal{F}$ to evaluate segmentation accuracy. 

\subsubsection{Results}
The proposed solution achieves 1st place on the complex video object segmentation track of the PVUW Challenge 2024. 
In the five submissions, we find that some inference parameters influence the performance, which are the test size, the memory interval, memory or not, the flip augmentation, and multi-scale fusion.

In a conclusion, we propose a robust solution for the task of video object segmentation, which helps the model understand the semantic information of the targets and generate discriminative queries of the target. 
In the end, we achieve 1st place on the complex video object segmentation track of the PVUW Challenge 2024 with 84.45\% $\mathcal{J}\&\mathcal{F}$.
The detailed version is under peer review.
The code and full version will be released as soon as possible.

\subsection{Yao\_Xu\_MTLab team} \label{sec:yao_xu}
\teamtitle{2nd Place Solution for MOSE Track in CVPR 2024 PVUW Workshop: Complex Video Object Segmentation \cite{xu20242nd}}
\teammembers{Zhensong Xu\textsuperscript{1}, Jiangtao Yao\textsuperscript{1}, Chengjing Wu\textsuperscript{1}, Ting Liu\textsuperscript{1}, and Luoqi Liu\textsuperscript{1}}
\teamaffiliations{MT Lab, Meitu Inc}

As illustrated in \cref{fig:1}, our solution takes Cutie as the baseline model. Then, we use instance segmentation and motion blur to augment the training data. Finally, during the inference stage, we employ TTA and memory strategy to improve the results. Details of the solution are described as follows.

\begin{figure}[t]
  \centering
  \includegraphics[width=0.95\linewidth]{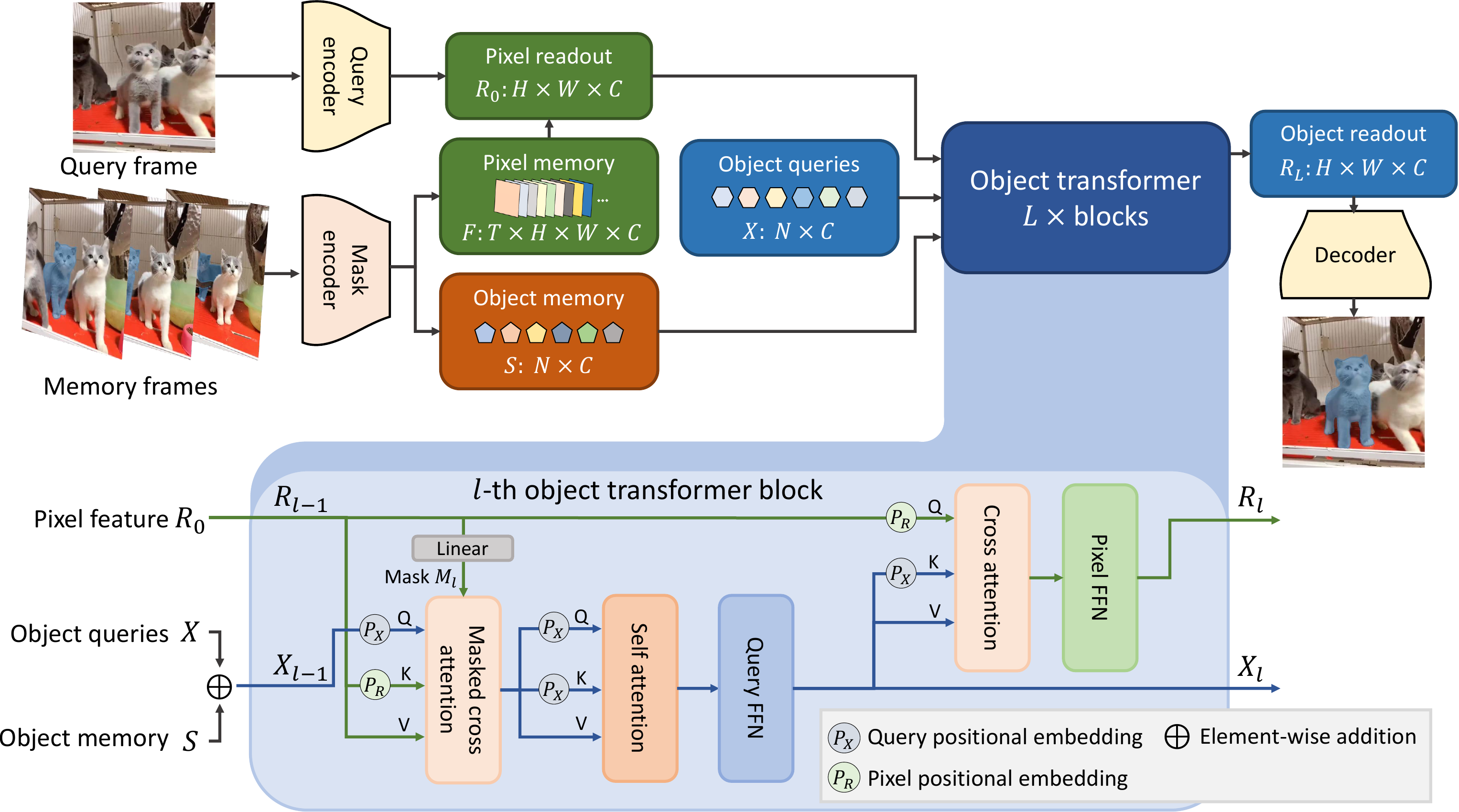}
   \caption{Architecture of Cutie\cite{Cutie}.}
   \label{fig:2}
\end{figure}
\subsubsection{Baseline model}
To ensure good performance under challenges such as frequent disappearance-reappearance, heavy occlusions, small and similar objects, we introduce Cutie as the strong baseline model, as shown in \cref{fig:2}. Cutie stores a high-resolution pixel memory $F$ and a high-level object memory $S$. The pixel memory is encoded from the memory frames and corresponding segmented masks. The object memory compresses object-level features from the memory frames. When a new query frame comes, it bidirectionally interacts with the object memory in a couple of object transformer blocks. Specifically, given the feature map of the query frame, the pixel readout $R_0$ is extracted by reading from the pixel memory with a sensory memory\cite{Cutie}, then the pixel readout interacts with the object memory and a set of learnable object queries through bottom-up foreground-background masked cross attention. Next, the obtained high-level object query representation communicates back with the pixel readout through top-down cross attention. The output pixel readout $R_l$ and object queries $X_l$ are sent to the next object transformer block. The final pixel readout will be combined with multi-scale features passed from skip connections for computing the output mask in the decoder. Cutie enriches pixel features with object-level semantics in a bidirectional fashion, hence is more robust to distractions such as occlusion and disappearance.

\subsubsection{Data augmentation}
Like most state-of-the-art VOS methods, Cutie also adopts a two-stage training paradigm. The first stage pretraining uses short video sequences generated from static images. Then main training is performed using VOS datasets in the second stage. However, the original Cutie fails to perform well when similar objects move in close proximity or suffers from serious motion blur. 

\begin{figure}[t]
    \centering
    \begin{tabular}{c@{\hspace{1pt}}c@{\hspace{1pt}}c@{\hspace{1pt}}c}
    \includegraphics[width=0.32\linewidth, height=0.20\linewidth]{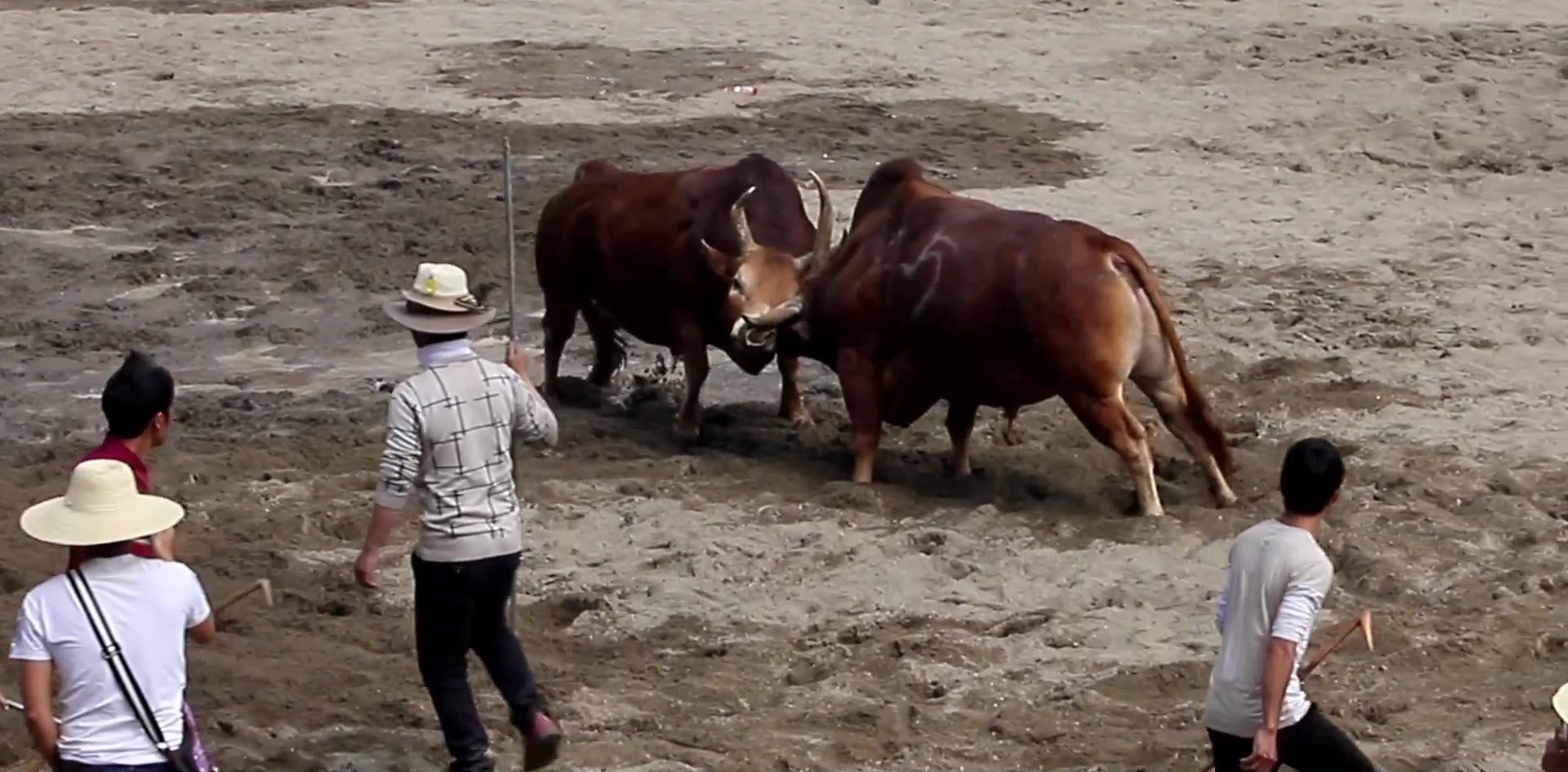} & 
        \includegraphics[width=0.32\linewidth, height=0.20\linewidth]{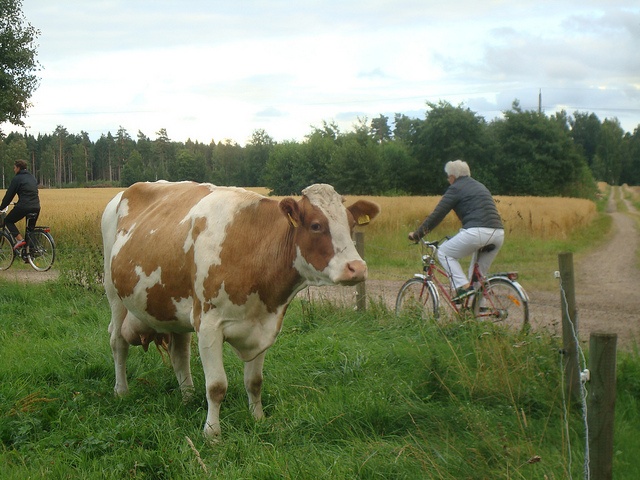} & 
        \includegraphics[width=0.32\linewidth, height=0.20\linewidth]{Figs/Yao_Xu_MTLab/Fig_3/bike.jpg} \\
        \includegraphics[width=0.32\linewidth, height=0.20\linewidth]{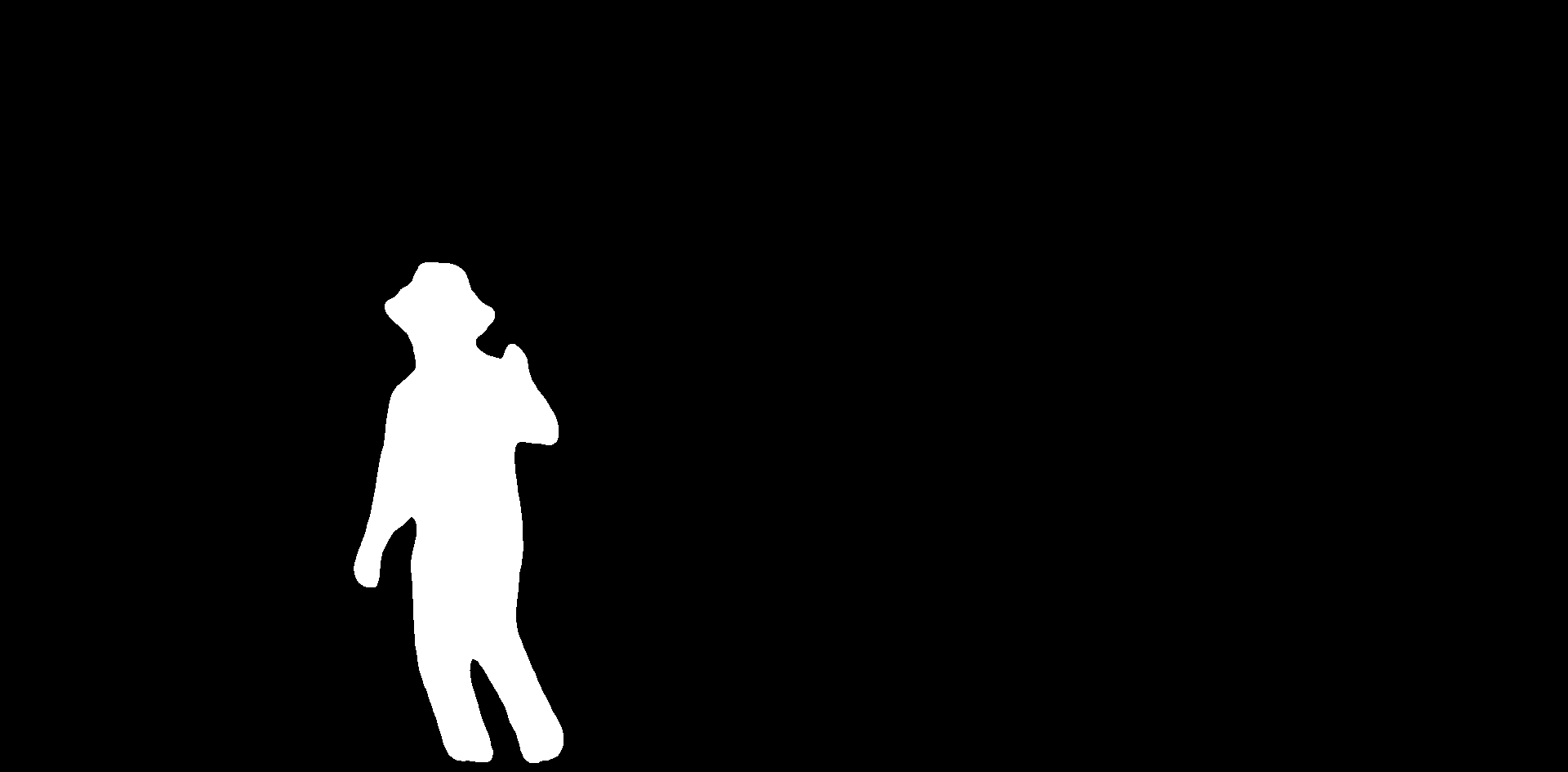} & 
        \includegraphics[width=0.32\linewidth, height=0.20\linewidth]{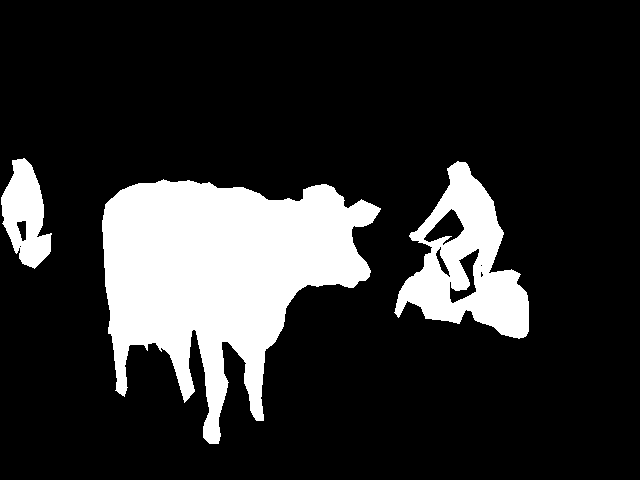} & 
        \includegraphics[width=0.32\linewidth, height=0.20\linewidth]{Figs/Yao_Xu_MTLab/Fig_3/bike_blur.jpg} \\
     \end{tabular}
     \vspace{-3mm}
    \caption{Examples of generated pretraining data and motion blur. Left: binary mask generated from the valid set and test set of MOSE. Middle: binary mask generated from COCO, the masks of different classes are merged into one mask. Right: example of motion blur in the horizontal direction.}
    \label{fig:3}
\end{figure}

To solve the above problems, we conduct data augmentation to enhance the training of Cutie. First, we employ the universal image segmentation model Mask2Former\cite{Mask2Former} to segment instance targets from the valid set and test set of MOSE. As shown in the left column of \cref{fig:3}, the segmented small objects represent typical object appearances in MOSE, which is helpful for learning the semantics of diverse objects in advance. Meanwhile, as shown in the middle column of \cref{fig:3}, we convert the instance annotations of COCO\cite{Coco} into independent binary masks. Here we select object classes such as human, animal and vehicle that frequently occur in MOSE to reduce  discrepancy between two data distributions. The acquired data is used as extra pretraining data to enable more robust semantics and improve discrimination ability against diverse objects of MOSE. Second, with the observation that motion blur is a significant challenge, we add motion blur with random kernel sizes and angles to both the pretraining and main training stages. An example of motion blur is shown in the right column of \cref{fig:3}. The proposed data augmentation aims at training towards better robustness and generalization.

\subsubsection{Inference time operations}
\paragraph{TTA.} We use two kinds of TTA: flipping and multi-scale data enhancement. We only conduct horizontal flipping since experiments show flipping in other directions is detrimental to performance. In addition, we inference results on the test set under three maximum shorter side resolutions: 600p, 720p and 800p. The multi-scale results are then averaged to get the final result.\\
\textbf{Memory strategy.} We find in experiments that larger memory banks and shorter memory intervals lead to better performance. Therefore, we adjust the maximum memory frames $T_{\max}$ to 18 and the memory interval to 1. 

\subsection{ISS team} \label{sec:ISS}
\teamtitle{3rd Place Solution for MOSE Track in CVPR 2024 PVUW workshop: Complex Video Object Segmentation \cite{liu20243rd}}
\teammembers{Xinyu Liu\textsuperscript{1},  Jing Zhang\textsuperscript{1},  Kexin Zhang\textsuperscript{1}, Yuting Yang\textsuperscript{1}, Licheng Jiao\textsuperscript{1}, Shuyuan Yang\textsuperscript{1},}
\teamaffiliations{Intelligent Perception and Image Understanding Lab, Xidian University}

\begin{figure*}[t]
\centering
\includegraphics[width=0.56\linewidth]{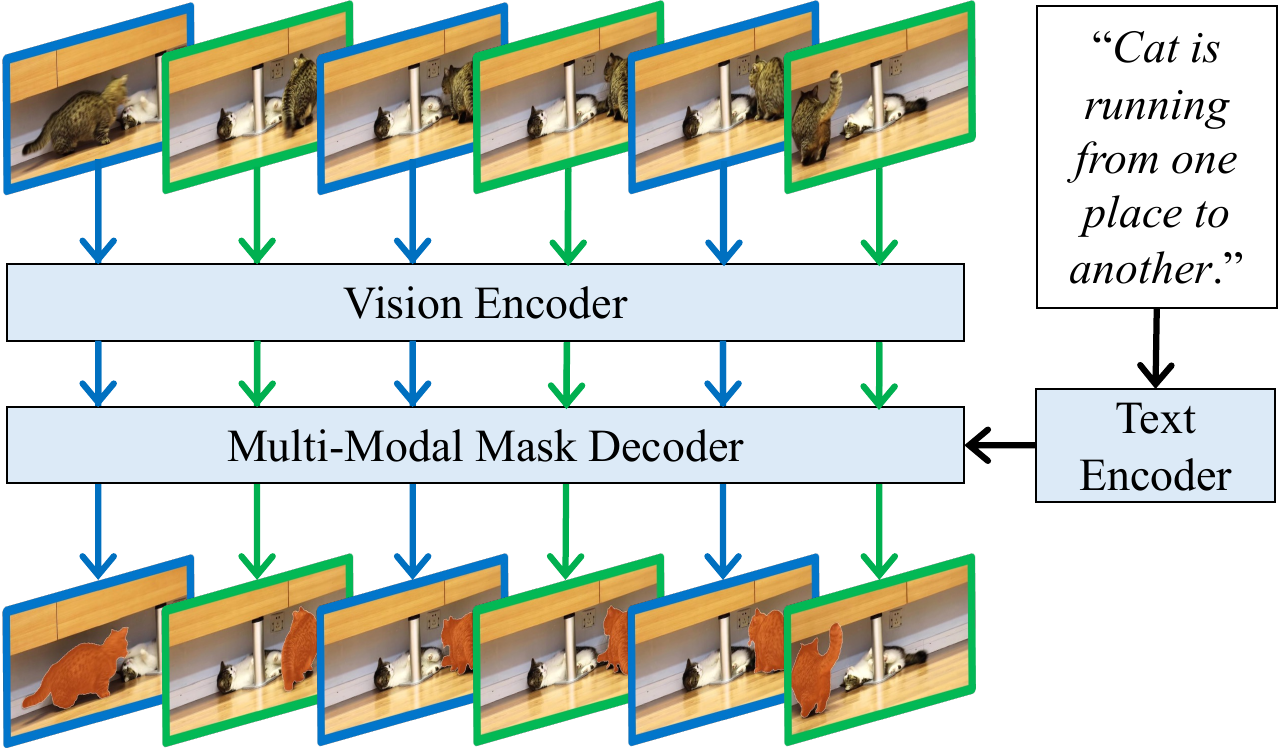}
   \caption{Overall framework of Tapall.ai team method, 1st place solution for MeViS Challenge in CVPR 2024. Given an input video, we divide all frames into $N$ subsets via non-continuous sampling. Here we take two subsets as an example. They are marked with \textcolor[RGB]{0,112,192}{Blue} and \textcolor[RGB]{0,176,80}{Green} boxes. Each subset is segmented individually, guided by the input text, and combined for the final results. }
\label{fig:tapall}
\end{figure*}

\subsubsection{Model}
Our approach is inspired by recent work on video object segmentation, particularly the Cutie framework,as shown in \cref{fig:2}. Cutie operates in a semi-supervised video object segmentation (VOS) setting, where it takes a first-frame segmentation as input and processes subsequent frames sequentially.
Cutie encodes segmented frames into a high-resolution pixel memory \(F\) and a high-level object memory \(S\). These memories are used for segmenting future frames. When segmenting a new frame, Cutie first retrieves an initial pixel readout \(R_0\) from the pixel memory using the encoded query features. This initial readout is typically noisy due to low-level pixel matching.

To enhance this initial readout, Cutie enriches \(R_0\) with object-level semantics using information from the object memory \(S\) and object queries \(X\). This is done through an object transformer with multiple transformer blocks. The final enriched output, \(R_L\), is then passed to the decoder to generate the output mask.
Cutie introduces three main contributions: object-transformer, sec:masked-attention, and object-memory.
The `Cutie-base' model is based on the `base' variant, utilizing ResNet-50 as the query encoder backbone. It consists of $C=256$ channels, $L=3$ object transformer blocks, and $N=16$ object queries.
The query and mask encoders are designed using ResNets. Following previous studies, we discard the final convolutional stage and employ the stride 16 feature. 

The object transformer block integrates both query FFN and pixel FFN components. The query FFN comprises a 2-layer MLP with a hidden size of $8C=2048$. Meanwhile, the pixel FFN utilizes two $3\times3$ convolutions with a reduced hidden size of $C=256$ to minimize computational overhead. The ReLU activation function is employed throughout the network.

\subsubsection{Inference}
When testing, the input video is upscaled to a resolution of 720p, which provides a higher density of pixel information compared to lower resolutions such as 480p.

In the context of the memory frame encoding, we update both the pixel memory and the object memory every $r$-th frame. The default value of $r$ is set to 3, following the same configuration used in the XMem framework. For subsequent memory frames, we employ a First-In-First-Out (FIFO) strategy, which ensures that the most recent information is retained while older data is gradually phased out. The choice of a predefined limit of $T_{\max} = 15$ for the total number of memory frames is a practical compromise.  Maintaining a history of 15 frames is generally adequate for effectively exploiting temporal correlations in VOS tasks. 

Based on these observations, we propose filtering affinities to retain only the top-$k$ entries. To further manage the memory capacity, we apply top-$k$ filtering with $k = 60$ to the pixel memory. Setting top-$k$ to 60 has the effect of prioritizing the most relevant pixel memories based on their attention scores, which is crucial for maintaining accurate segmentation over time while preventing the memory from being overwhelmed with less significant information. 

In the final testing phase, we employed filpping Test-Time Augmentation (TTA), which is a strategy that enhances the robustness and accuracy of predictions by incorporating a variety of augmented versions of the input data.

\begin{table}[t]
   \renewcommand\arraystretch{1.2}
   \centering
   \caption{\textbf{MeViS Challenge results and final rankings.}}
   \vspace{-3mm}
   \setlength{\tabcolsep}{3.6mm}{\begin{tabular}{clrrr}
      \thickhline
\rowcolor{gray!60}Rank & Team  & \multicolumn{1}{c}{$\mathcal{J}$} & \multicolumn{1}{c}{$\mathcal{F}$} & \multicolumn{1}{c}{$\mathcal{J\&F}$} \\
     \hline
     1     & Tapallai & 50.5  & 58.5  & 54.5 \\
     \rowcolor{gray!20}2     & CASIA\_IVA & 51.0  & 57.4  & 54.2 \\
     3     & TIME  & 46.1  & 56.9  & 51.5 \\
     \rowcolor{gray!20}4     & Phan  & 45.6  & 55.9  & 50.8 \\
     5     & LIULINKAI & 39.3  & 46.1  & 42.7 \\
     \rowcolor{gray!20}-     & (Baseline) & 34.1  & 39.9  & 37.0 \\
     \thickhline
   \end{tabular}}%
   \label{tab:results_MeViS}%
 \end{table}%

\section{MeViS Challenge Methods and Teams}

For MeViS~\cite{MeViS}, out of 225 teams joined the competition in total, 50 teams participated the valid phase and 5 teams entered the challenge phase. The final results of the MeViS Track are reported in \Cref{tab:results_MeViS}.

\noindent\href{https://henghuiding.github.io/MeViS/ChallengeCVPR2024}{https://henghuiding.github.io/MeViS/ChallengeCVPR2024}

\subsection{Tapall.ai team} \label{sec:tapall}
\teamtitle{1st Place Solution for MeViS Track in CVPR 2024 PVUW Workshop: Motion Expression guided Video Segmentation~\cite{gao20241st}}
\teammembers{Mingqi Gao\textsuperscript{1,2,4}, Jingnan Luo\textsuperscript{2}, Jinyu Yang\textsuperscript{1}, Jungong Han\textsuperscript{3,4}, Feng Zheng\textsuperscript{1,2}}
\teamaffiliations{\textsuperscript{1}Tapall.ai\\
\textsuperscript{2}Southern University of Science and Technology\\
\textsuperscript{3}University of Sheffield\\
\textsuperscript{4}University of Warwick}

\noindent \textbf{Method:} Our solution explores the value of static-dominant data and frame sampling for the challenging MeViS benchmark. As shown in Fig.~\ref{fig:tapall}, we consider MUTR \cite{yan2024referred} as the baseline architecture. With pre-trained parameters on the Ref-COCO series \cite{yu2016modeling,mao2016generation} and Ref-YouTube-VOS~\cite{seo2020urvos}, we fine-tune them on MeViS. Masks with one-to-more text-object pairs are considered as a whole to encourage adaptive object perception based on texts. To balance comprehensive understanding and efficiency, we split long input videos into sub-videos via frame sampling. With these improvements, our solution ranks 1st (54.5 $\mathcal{J}\&\mathcal{F}$) in the MeViS Track. 

\begin{figure*}[t]
\begin{center}
\includegraphics[width=1\linewidth]{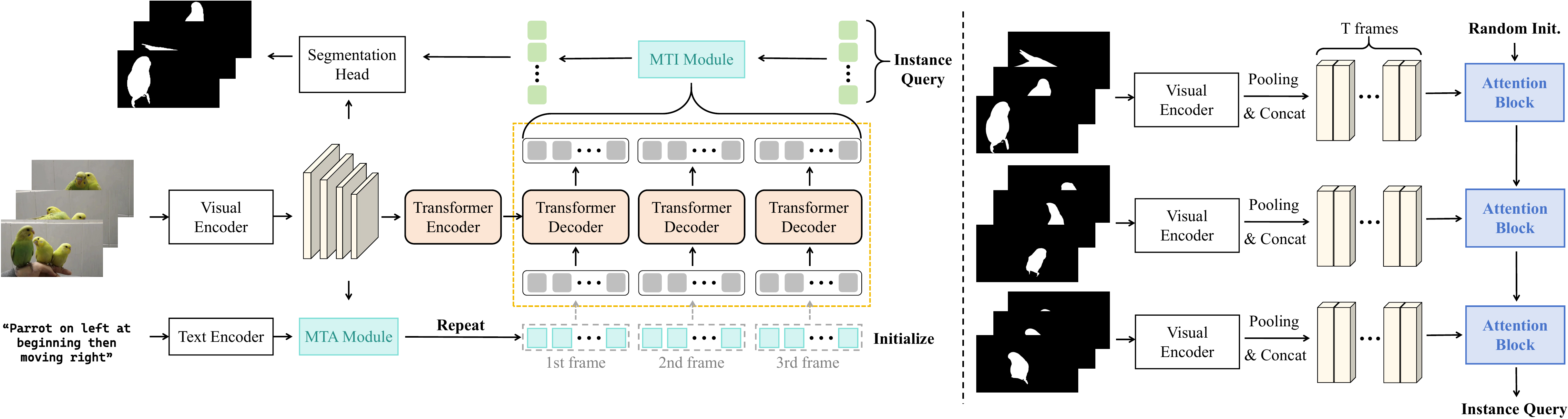}
\end{center}
   \vspace{-5mm}
\caption{Overall framework of CASIA\_IVA team method, 2nd place solution for MeViS Challenge in CVPR 2024.}
\label{fig:casia_iva}
\end{figure*}

Experiments on the MeViS valid set (48.6 $\mathcal{J}\&\mathcal{F}$) indicate that the static-dominant data still contribute to this challenging setting due to their sufficient and well-aligned object masks and texts. In addition, ablations on sampling schemes reveal that there is much room for improvement in temporal modelling over long videos. Limited by computational resources, the temporal modules are trained with pseudo videos with less frames. During inference, however, videos have more temporal contexts. This inconsistency leads to considering fewer frames (sampled) in temporal modules outperform the one with all frames. We hope these findings are helpful for future research.

\subsection{CASIA\_IVA team} \label{sec:casia}
\teamtitle{2nd Place Solution for MeViS Track in CVPR 2024 PVUW Workshop: Motion Expression guided Video Segmentation~\cite{cao20242nd}}
\teammembers{Bin Cao\textsuperscript{1,2,3}, Yisi Zhang\textsuperscript{4}, Xuanxu Lin\textsuperscript{2}, Xingjian He\textsuperscript{1}, Bo Zhao\textsuperscript{3}, Jing Liu\textsuperscript{1,2}}
\teamaffiliations{\textsuperscript{1}Institute of Automation, Chinese Academy of Sciences \\
                \textsuperscript{2}School of Artificial Intelligence, University of Chinese Academy of Sciences \\
                \textsuperscript{3}Beijing Academy of Artificial Intelligence \\
                \textsuperscript{4}University of Science and Technology Beijing 
                }

\noindent \textbf{Method:} As shown in \cref{fig:casia_iva} We attempt to introduce instance information to mitigate the issue of inconsistent predicted results across multiple frames. Specifically, we employ a video instance segmentation model to extract all instance masks in the video. Next, we utilize a query with random initialization to aggregate all instances information through our designed attention-based block including a cross-attention layer and a set of self-attention layers, FFN layers. We employ MUTR\cite{yan2024referred} as our basic model and utilize the query with instance information for query initialization. Most previous work in RVOS sample frames around a center point, allowing model to process part of video. In our solution, we sample frames in a manner of global sampling. We divide the entire video into a few phases and sample one frame in every phase to obtain a video clip. To further improve performance, we employ HQ-SAM\cite{ke2024segment} with VIT-L backbone utilizing default parameters for spatial refinement. Thanks to the superior performance of DVIS\cite{zhang2023dvis}, MUTR and HQ-SAM, our solution achieves a score of 49.92 \( \mathcal{J} \)\&\( \mathcal{F} \) on the MeViS validation set and 54.20 \( \mathcal{J} \)\&\( \mathcal{F} \) on the MeViS test set, ranking 2nd Place for MeViS Track in CVPR 2024 PVUW Workshop: Motion Expression guided Video Segmentation.

\subsection{TIME team} \label{sec:time}
\teamtitle{3rd Place Solution for MeViS Track in CVPR 2024 PVUW Workshop: Motion Expression guided Video Segmentation~\cite{pan20243rd}}
\teammembers{Feiyu Pan\textsuperscript{1}, Hao Fang\textsuperscript{1}, and Xiankai Lu\textsuperscript{1}}
\teamaffiliations{\textsuperscript{1}School of Software, Shandong University}

\begin{figure*}[h]
\begin{center}
\includegraphics[width=0.9\linewidth]{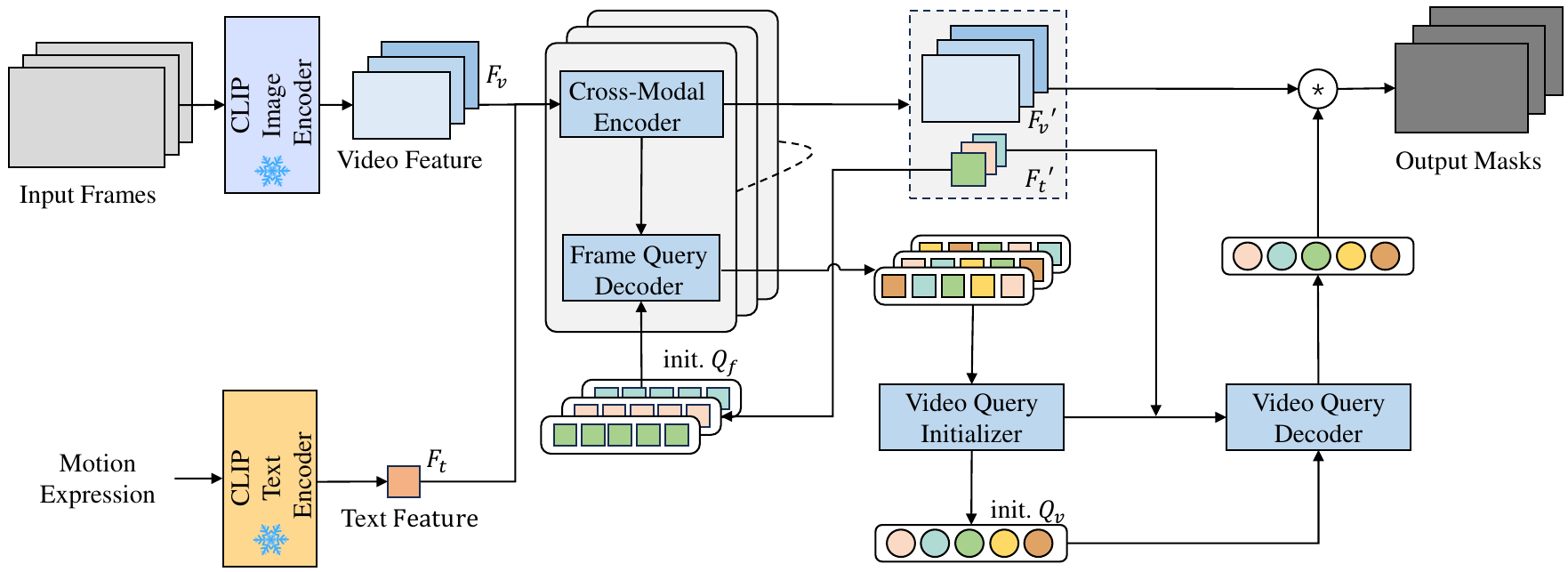}
\end{center}
   \vspace{-3mm}
\caption{Overall framework of TIME team method, 3rd place solution for MeViS Challenge in CVPR 2024.}
\label{fig:time}
\end{figure*}

As shown in Fig.~\ref{fig:time}, we propose using frozen pre-trained vision-language models (VLM) as backbones, with a specific emphasis on enhancing cross-modal feature interaction. Firstly, we use frozen convolutional CLIP~\cite{ConvNext,CLIP} backbone to generate feature-aligned vision and text features. We do not fine tune the CLIP backbone to preserve pre-trained knowledge of vision-language association. This not only alleviates the issue of domain gap, but also greatly reduces training costs. Secondly, we add more cross-modal feature fusion in the pipeline to enhance the utilization of multi-modal information. We design three cross-modal feature interaction module in the model, including cross-modal encoder, frame query decoder and video query decoder. These modules enhance video and text features through simple cross-attention. Furthermore, inspired by LBVQ~\cite{LBVQ}, we propose a novel video query initialization method to generate higher quality video queries. Specifically, we perform bipartite matching and reorder frame queries, then aggregate them in a weighted manner to initialize video queries. Without using any additional training data, our method~\cite{pan20243rd} achieved 46.9 \( \mathcal{J} \)\&\( \mathcal{F} \) on the MeViS val set, 51.5 \( \mathcal{J} \)\&\( \mathcal{F} \) on the MeViS test set and ranked 3rd place for MeViS Track in CVPR 2024 PVUW workshop: Motion Expression guided Video Segmentation.

\subsubsection{Cross-modal Encoder}
Given an \texttt{(Video, Text)} pair, we extract multi-frame multi-scale image features $F_v$ with CLIP image encoder, and text features $F_t$ with CLIP test encoder. Due to the use of convolutional CLIP image encoder~\cite{ConvNext,CLIP}, we can extract multi-scale features from the outputs of different blocks. After extracting vanilla video and text features, we fed them into a cross-modal encoder for cross-modal feature fusion. The cross-modal encoder is built on top of the pixel decoder of Mask2Former~\cite{Mask2Former}, which leverages the Deformable self-attention to enhance image features. We add an image-to-text cross-attention and a text-to-image cross-attention for feature fusion. These modules help align features of different modalities, ultimately obtaining enhanced image features $F^{'}_{v}$ and text features $F^{'}_{t}$.

\subsubsection{Frame Query Decoder}
We develop a frame query decoder to independently generate frame queries $Q_{f} \in \mathbb{R}^{T \times N_{f}\times C}$ for each frame. Frame queries are directly initialized by text features, then are fed into a text cross-attention layer to combine text features, an image cross-attention layer to combine image features, a self-attention layer, and an FFN layer in each frame query decoder layer. Each decoder layer has an extra text cross-attention layer compared with the transformer decoder layer of Mask2Former~\cite{Mask2Former}, as we need to inject text information into queries for better modality alignment.

\subsubsection{Video Query Initializer}
After generating frame-level representation, the next step is to generate video queries  $Q_{v} \in \mathbb{R}^{N_{v}\times C}$ to represent the entire video clip. Inspired by LBVQ~\cite{LBVQ}, video queries have great similarity to frame queries per frame, and their essence is the fusion of frame queries. Instead of the simple text feature initialization strategy, we aggregate frame queries to achieve video query initialization. Firstly, the Hungarian matching algorithm is utilized to match the $Q_f$ of adjacent frames. The purpose of this operation is to ensure that the instance order of each frame query is consistent. Then, due to the varying importance of each frame, we aggregate frame queries using learnable weights. The weights of different frames are maintained as a sum of 1 through the \text{Softmax} function.

\subsubsection{Video Query Decoder}
After obtaining the initialized video queries, they are fed into the video query decoder for layer by layer refinement. Video queries are fed into a text cross-attention layer to combine text features, an query cross-attention layer to combine frame queries features, a self-attention layer, and an FFN layer in each video query decoder layer. The video queries of the last layer will be dot multiplied with image features to generate the final mask.

\section{Conclusion \& Future Work}

This paper summarizes the methods and results of PVUW 2024 challenge on complex video understanding, including MOSE challenge and MeViS challenge. In the challenges, we have seen a significant improvement in performance. 
In the MOSE challenge, most works focus on using memory to preserve long-term video perception. In the MeViS challenge, there is a growing interest in modeling language with temporal relationships within videos. Despite these advancements, qualitative results indicate that accurately predicting masks remains a challenge.
Looking ahead, it is promising to consider integrating SAM (Segment Anything Model~\cite{SAM}) and Large Language Models (LLM) in future challenges. We hope that the MOSE and MeViS challenges will continue to attract new researchers and participants to the field of complex video understanding.
{\small
\bibliographystyle{ieee_fullname}
\bibliography{egbib}

\begin{thebibliography}{10}\itemsep=-1pt

\bibitem{cao20242nd}
Bin Cao, Yisi Zhang, Xuanxu Lin, Xingjian He, Bo Zhao, and Jing Liu.
\newblock 2nd place solution for mevis track in cvpr 2024 pvuw workshop: Motion expression guided video segmentation.
\newblock {\em arXiv preprint arXiv:2406.13939}, 2024.

\bibitem{Mask2Former}
Bowen Cheng, Ishan Misra, Alexander~G Schwing, Alexander Kirillov, and Rohit Girdhar.
\newblock Masked-attention mask transformer for universal image segmentation.
\newblock In {\em CVPR}, 2022.

\bibitem{Coco}
Ho~Kei Cheng, Seoung~Wug Oh, Brian Price, Joon-Young Lee, and Alexander Schwing.
\newblock Putting the object back into video object segmentation.
\newblock {\em arXiv preprint arXiv:2310.12982}, 2023.

\bibitem{Cutie}
Ho~Kei Cheng and Alexander~G Schwing.
\newblock Xmem: Long-term video object segmentation with an atkinson-shiffrin memory model.
\newblock In {\em ECCV}, 2022.

\bibitem{ding2020phraseclick}
Henghui Ding, Scott Cohen, Brian Price, and Xudong Jiang.
\newblock Phraseclick: toward achieving flexible interactive segmentation by phrase and click.
\newblock In {\em ECCV}, 2020.

\bibitem{CCL}
Henghui Ding, Xudong Jiang, Bing Shuai, Ai~Qun Liu, and Gang Wang.
\newblock Context contrasted feature and gated multi-scale aggregation for scene segmentation.
\newblock In {\em CVPR}, 2018.

\bibitem{MeViS}
Henghui Ding, Chang Liu, Shuting He, Xudong Jiang, and Chen~Change Loy.
\newblock {MeViS}: A large-scale benchmark for video segmentation with motion expressions.
\newblock In {\em ICCV}, 2023.

\bibitem{MOSE}
Henghui Ding, Chang Liu, Shuting He, Xudong Jiang, Philip~HS Torr, and Song Bai.
\newblock {MOSE}: A new dataset for video object segmentation in complex scenes.
\newblock In {\em ICCV}, 2023.

\bibitem{ding2021vision}
Henghui Ding, Chang Liu, Suchen Wang, and Xudong Jiang.
\newblock Vision-language transformer and query generation for referring segmentation.
\newblock In {\em ICCV}, 2021.

\bibitem{VLTPAMI}
Henghui Ding, Chang Liu, Suchen Wang, and Xudong Jiang.
\newblock {VLT}: Vision-language transformer and query generation for referring segmentation.
\newblock {\em IEEE TPAMI}, 2023.

\bibitem{LBVQ}
Hao Fang, Tong Zhang, Xiaofei Zhou, and Xinxin Zhang.
\newblock Learning better video query with sam for video instance segmentation.
\newblock {\em IEEE TCSVT}, 2024.

\bibitem{gao20241st}
Mingqi Gao, Jingnan Luo, Jinyu Yang, Jungong Han, and Feng Zheng.
\newblock 1st place solution for mevis track in cvpr 2024 pvuw workshop: Motion expression guided video segmentation.
\newblock {\em arXiv preprint arXiv:2406.07043}, 2024.

\bibitem{DsHmp}
Shuting He and Henghui Ding.
\newblock Decoupling static and hierarchical motion perception for referring video segmentation.
\newblock In {\em CVPR}, 2024.

\bibitem{ke2024segment}
Lei Ke, Mingqiao Ye, Martin Danelljan, Yu-Wing Tai, Chi-Keung Tang, Fisher Yu, et~al.
\newblock Segment anything in high quality.
\newblock In {\em NeurIPS}, 2024.

\bibitem{SAM}
Alexander Kirillov, Eric Mintun, Nikhila Ravi, Hanzi Mao, Chloe Rolland, Laura Gustafson, Tete Xiao, Spencer Whitehead, Alexander~C Berg, Wan-Yen Lo, et~al.
\newblock Segment anything.
\newblock In {\em ICCV}, 2023.

\bibitem{li2023transformer}
Xiangtai Li, Henghui Ding, Wenwei Zhang, Haobo Yuan, Jiangmiao Pang, Guangliang Cheng, Kai Chen, Ziwei Liu, and Chen~Change Loy.
\newblock Transformer-based visual segmentation: A survey.
\newblock {\em arXiv preprint arXiv:2304.09854}, 2023.

\bibitem{liu2023gres}
Chang Liu, Henghui Ding, and Xudong Jiang.
\newblock {GRES:} generalized referring expression segmentation.
\newblock In {\em CVPR}, pages 23592--23601, 2023.

\bibitem{MA3Net}
Chang Liu, Henghui Ding, Yulun Zhang, and Xudong Jiang.
\newblock Multi-modal mutual attention and iterative interaction for referring image segmentation.
\newblock {\em IEEE TIP}, 2023.

\bibitem{ISFP}
Chang Liu, Xudong Jiang, and Henghui Ding.
\newblock Instance-specific feature propagation for referring segmentation.
\newblock {\em IEEE TMM}, 2023.

\bibitem{liu20243rd}
Xinyu Liu, Jing Zhang, Kexin Zhang, Yuting Yang, Licheng Jiao, and Shuyuan Yang.
\newblock 3rd place solution for mose track in cvpr 2024 pvuw workshop: Complex video object segmentation.
\newblock {\em arXiv preprint arXiv:2406.03668}, 2024.

\bibitem{ConvNext}
Zhuang Liu, Hanzi Mao, Chao-Yuan Wu, Christoph Feichtenhofer, Trevor Darrell, and Saining Xie.
\newblock A convnet for the 2020s.
\newblock In {\em CVPR}, 2022.

\bibitem{mao2016generation}
Junhua Mao, Jonathan Huang, Alexander Toshev, Oana Camburu, Alan~L Yuille, and Kevin Murphy.
\newblock Generation and comprehension of unambiguous object descriptions.
\newblock In {\em CVPR}, pages 11--20, 2016.

\bibitem{miao1st}
Deshui Miao, Xin Li, Zhenyu He, Yaowei Wang, and Ming-Hsuan Yang.
\newblock 1st place solution for mose track in cvpr 2024 pvuw workshop: Complex video object segmentation.
\newblock {\em arXiv preprint arXiv:2406.04600}, 2024.

\bibitem{pan20243rd}
Feiyu Pan, Hao Fang, and Xiankai Lu.
\newblock 3rd place solution for mevis track in cvpr 2024 pvuw workshop: Motion expression guided video segmentation.
\newblock {\em arXiv preprint arXiv:2406.04842}, 2024.

\bibitem{codalab_competitions_JMLR}
Adrien Pavao, Isabelle Guyon, Anne-Catherine Letournel, Dinh-Tuan Tran, Xavier Baro, Hugo~Jair Escalante, Sergio Escalera, Tyler Thomas, and Zhen Xu.
\newblock Codalab competitions: An open source platform to organize scientific challenges.
\newblock {\em Journal of Machine Learning Research}, 2023.

\bibitem{Perazzi2016}
F. Perazzi, J. Pont-Tuset, B. McWilliams, L. {Van Gool}, M. Gross, and A. Sorkine-Hornung.
\newblock A benchmark dataset and evaluation methodology for video object segmentation.
\newblock In {\em CVPR}, 2016.

\bibitem{CLIP}
Alec Radford, Jong~Wook Kim, Chris Hallacy, Aditya Ramesh, Gabriel Goh, Sandhini Agarwal, Girish Sastry, Amanda Askell, Pamela Mishkin, Jack Clark, et~al.
\newblock Learning transferable visual models from natural language supervision.
\newblock In {\em ICML}, 2021.

\bibitem{seo2020urvos}
Seonguk Seo, Joon-Young Lee, and Bohyung Han.
\newblock Urvos: Unified referring video object segmentation network with a large-scale benchmark.
\newblock In {\em ECCV}, 2020.

\bibitem{wu2024towards}
Jianzong Wu, Xiangtai Li, Shilin Xu, Haobo Yuan, Henghui Ding, Yibo Yang, Xia Li, Jiangning Zhang, Yunhai Tong, Xudong Jiang, et~al.
\newblock Towards open vocabulary learning: A survey.
\newblock {\em IEEE TPAMI}, 2024.

\bibitem{vos2018}
Ning Xu, Linjie Yang, Yuchen Fan, Dingcheng Yue, Yuchen Liang, Jianchao Yang, and Thomas~S. Huang.
\newblock Youtube-vos: {A} large-scale video object segmentation benchmark.
\newblock {\em CoRR}, abs/1809.03327, 2018.

\bibitem{xu20242nd}
Zhensong Xu, Jiangtao Yao, Chengjing Wu, Ting Liu, and Luoqi Liu.
\newblock 2nd place solution for mose track in cvpr 2024 pvuw workshop: Complex video object segmentation.
\newblock {\em arXiv preprint arXiv:2406.08192}, 2024.

\bibitem{yan2024referred}
Shilin Yan, Renrui Zhang, Ziyu Guo, Wenchao Chen, Wei Zhang, Hongyang Li, Yu Qiao, Hao Dong, Zhongjiang He, and Peng Gao.
\newblock Referred by multi-modality: A unified temporal transformer for video object segmentation.
\newblock In {\em AAAI}, 2024.

\bibitem{yu2016modeling}
Licheng Yu, Patrick Poirson, Shan Yang, Alexander~C Berg, and Tamara~L Berg.
\newblock Modeling context in referring expressions.
\newblock In {\em ECCV}, 2016.

\bibitem{zhang2023dvis}
Tao Zhang, Xingye Tian, Yu Wu, Shunping Ji, Xuebo Wang, Yuan Zhang, and Pengfei Wan.
\newblock Dvis: Decoupled video instance segmentation framework.
\newblock In {\em ICCV}, 2023.

\end{thebibliography}
}

\end{document}